\documentclass{article}

\usepackage[T1]{fontenc}
\usepackage{lmodern}
\usepackage{graphicx}
\usepackage{amsmath}
\usepackage{amsfonts}
\usepackage{fancyvrb}
\usepackage{fvextra}
\usepackage{booktabs}

\usepackage{float}

\usepackage{graphicx}
\usepackage{subcaption}
\usepackage{booktabs}
\usepackage{tabularx}
\usepackage{wrapfig}

\usepackage[preprint]{corl_2026} 

\title{Do VLAs Know They Will Fail? Interpretable Success and Failure Patterns in Physical AI Systems}

\author{
  Christian A.~Rogers\thanks{Equal Contribution}\\
  Kahlert School of Computing\\
  University of Utah\\
  \texttt{Christian.Rogers@utah.edu} \\
  \And
  Qian Lin\footnotemark[1]\\
  Kahlert School of Computing\\
  University of Utah \\
  \texttt{Qian.Lin@utah.edu} \\
  \And
  Daniel S.~Brown \\
  Kahlert School of Computing\\
  University of Utah\\
  \texttt{Daniel.S.Brown@utah.edu} \\
}

\begin{document}
\maketitle

\begin{abstract}
    Vision-Language-Action (VLA) models have emerged as powerful general-purpose robot policies, but their successes and failures remain difficult to diagnose and understand. In this work, we show that sparse autoencoder (SAE) representations of OpenVLA activations contain outcome-predictive information and can be used to predict task failure before the robot reaches an unrecoverable state.
    Moreover, to provide insights into what common behaviors may contribute to these outcomes, we identify individual SAE features whose strong activations are associated with specific stages of success or failure, and uncover recurring behavioral patterns by localizing temporal intervals where these features exhibit significant activation changes.
    We further develop a VLM-based interpretation pipeline to generate behavioral descriptions of these patterns and validate their correspondence to feature activations on held-out intervals.
\end{abstract}



\section{Introduction}

    Vision-language-action (VLA) models have achieved strong generalist performance by scaling model capacity and training data~\cite{zitkovich2023rt2,kim2024openvla}, but their internal representations make it difficult to diagnose why a rollout succeeds or fails.
    Recent work shows that these representations anticipate failure. SAFE learns a multitask failure detector from internal VLA features and uses conformal prediction to trade detection accuracy against warning time~\cite{gu2025safe}. Mahato and Ren show that near-term failure is linearly decodable from OpenVLA activations under visual distribution shift~\cite{mahato2026early}. FPC-VLA instead predicts failures with an external VLM supervisor~\cite{yang2026fpcvla}.
    These works report warning time relative to the end of the episode. The problem is that every action an agent takes changes how easy the remaining task is to complete; thus, these approaches may trigger an alarm after the task has become unrecoverable. We therefore build on this line of research by measuring lead time against a \emph{point of no return}, a timestep after which the only reachable states by branched policies are failures. 

    More importantly, we aim to move beyond predicting outcomes toward identifying the behavioral patterns strongly associated with success or failure. Sparse autoencoders (SAEs) provide a promising approach by decomposing dense, polysemantic activations into sparse, potentially interpretable features~\cite{bricken2023towards,cunningham2023sparse}.
    Recent work has applied this technique to VLAs. Haon et al.~\cite{haon2025mechanistic} identify interpretable, steerable directions in OpenVLA. Swann et al.~\cite{swann2026sparse} find SAE features for motion and semantics, some of which causally steer behavior. Event-Grounded SAEs tie features to behaviorally salient events in closed-loop rollouts~\cite{jin2026eventgrounded}.
    These works characterize features by what they represent or control. We ask a complementary question: which sparse features track whether a rollout will succeed, and what behavior do they correspond to?

    We show that SAE features preserve the outcome-predictive information in OpenVLA activations, and that, evaluated against the point of no return, the predictor separates eventual failures from successes while the failures are still recoverable. We then select individual features by their association with trajectory outcomes and automatically localize the critical intervals where they emerge. Finally, we describe these intervals with a VLM and validate the descriptions on held-out clips via detection scoring~\cite{bills2023language,paulo2024automatically}.
    Together, these steps connect early failure warning to interpretable, failure-relevant behavior in VLA policies. An extended discussion of related work is in Appendix~\ref{sec:citations-extended}.


\section{Experimental Results}
\label{sec:result}

    \subsection{Problem Setup}

        We run OpenVLA-7B~\cite{kim2024openvla} fine-tuned on LIBERO-10~\cite{liu2023libero} greedily from the 50 benchmark initial states of each of the ten tasks, giving 500 rollouts that end either in success or in failure at a 520-step timeout (51.6\% success; 95 trajectories held out for evaluation).
        Let $\mathbf{h}_{i,t}$ denote the layer-0 residual-stream activation at timestep $t$ of trajectory $i$, averaged over image tokens. While dense activations may contain outcome-predictive information, their polysemanticity makes it difficult to identify the specific behavioral patterns associated with success or failure~\cite{bricken2023towards,cunningham2023sparse}. We therefore use a TopK sparse autoencoder (2{,}048 features, $K=64$) to decompose them into sparse feature activations $\mathbf{a}_{i,t}$.
        We ask whether $\mathbf{h}_{i,t}$ and $\mathbf{a}_{i,t}$ predict the eventual outcome, and whether individual SAE features reveal interpretable success- and failure-related behavior (task, splits, and SAE details are in Appendix~\ref{app:setup}).

    \subsection{Online Outcome Prediction}
    \label{sec:online-prediction}
    
        We first test whether the raw activations and SAE features contain information predictive of the eventual trajectory outcome. Specifically, at each timestep $t$, we train a logistic regressor $f_t$ using either the raw activations or SAE features to predict the final outcome of an ongoing trajectory, taking its predicted failure probability as $p_{\mathrm{fail}}(t)$. To reduce transient activation noise, we consider activations within a rolling window of $T=40$ steps rather than single-step activations. The predictor input is the rolling mean raw activation $m^{h}_{i,t}=1/T\sum_{s=t-T+1}^{t}\mathbf{h}_{i,s}$ or the rolling mean SAE representation $m^{a}_{i,t}=1/T\sum_{s=t-T+1}^{t}\mathbf{a}_{i,s}$.
        As successful trajectories tend to terminate earlier, the activations at large $t$ predominantly come from failed trajectories, causing severe label imbalance and making high accuracy easy to achieve. 
        We therefore compare our predictor with a majority baseline that takes no input and always predicts the majority outcome label at each timestep $t$.
        
        \begin{figure}[t]
        \vskip -0.2in
            \centering
            \begin{subfigure}[t]{0.31\linewidth}
                \includegraphics[width=\linewidth]{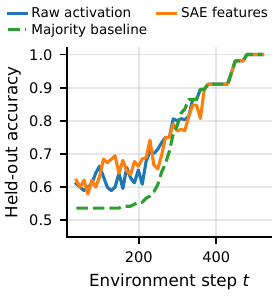}
                \caption{Held-out accuracy and majority baseline.}\label{fig:online-accuracy}
            \end{subfigure}\hfill
            \begin{subfigure}[t]{0.31\linewidth}
                \includegraphics[width=\linewidth]{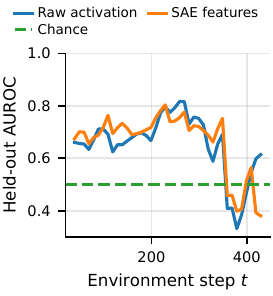}
                \caption{Held-out AUROC (steps with $\geq$5 trajectories per outcome).}\label{fig:online-auroc}
            \end{subfigure}\hfill
            \begin{subfigure}[t]{0.32\linewidth}
                \includegraphics[width=\linewidth]{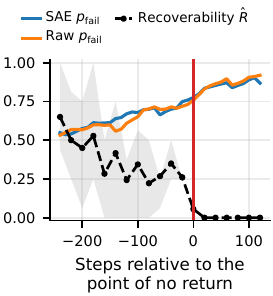}
                \caption{Failures aligned on their PNR ($n=$ 35).}\label{fig:aligned-pnr}
            \end{subfigure}
            \caption{
            For late timesteps, almost all trajectories are failures, so low $t$ performance is most relevant.
            (a) Raw activations and SAE features outperform
            the majority baseline at early timesteps.
            (b) Both representations achieve above-chance AUROC, indicating that they retain outcome-predictive information.
            (c) For held-out failures aligned at their point of no return (PNR, red line), predicted failure probability rises as estimated recoverability declines.
            }
            \label{fig:online-prediction}
            \vskip -0.2in
        \end{figure}

        Figure~\ref{fig:online-prediction}(a) shows that raw activations and SAE features achieve similar accuracy and outperform the majority baseline at early-to-mid timesteps. Their AUROCs are also nearly identical (0.80 and 0.81, respectively, over all held-out frames), remaining above chance from the first 80 steps (0.68) through steps 240-319 (0.75). These results indicate that SAE features preserve most of the outcome-predictive information in the original OpenVLA activations.

        \textbf{Recoverability-grounded early warning.} Predicting failure before an episode ends does not necessarily provide an actionable warning, as the robot may have already reached an unrecoverable state. To evaluate whether our predictor identifies failures while recovery is still possible, we estimate state recoverability by restoring saved simulator states and sampling policy continuations. We define the point of no return (PNR) as the earliest probe after which all subsequent probes have estimated recoverability $\hat R\le0.1$. An alarm is triggered when $p_{\mathrm{fail}}\ge\tau$ for two consecutive evaluations and is actionable if it occurs before the PNR.

        \begin{wraptable}[10]{R}{.43\linewidth} %
        \vspace{-.5cm}
            \centering
            \scriptsize
            \caption{Alarms relative to the point of no return (held-out). False alarms: fraction of successful trajectories that raise an alarm. Actionable: fraction of failures alarmed before their PNR. Lead: median steps from alarm to PNR for actionable alarms.}
            \label{tab:alarms}
            \begin{tabular}{rcccccc}
                \toprule
                \cmidrule(lr){1-4}
                $\tau$ & False alarms & Actionable & Lead  \\
                \midrule
                0.5 & 0.64 & 0.88 & 190 \\
                0.7 & 0.25 & 0.55 & 160 \\
                0.9 & 0.14 & 0.33 & 70  \\
                \bottomrule
            \end{tabular}
        \end{wraptable}
        
        \textbf{Early warning performance.} Figures~\ref{fig:online-prediction}(c) illustrate a trade-off between prediction accuracy and actionability: although accuracy generally increases over time, recoverability declines, leaving fewer opportunities for intervention. Before the estimated PNR, the SAE predictor distinguishes eventual failures from successes with an AUROC of 0.71 (95\% CI [0.62, 0.79]), compared with 0.65 [0.57, 0.74] for raw activations. Table~\ref{tab:alarms} reports the early warning performance under different alarm thresholds. As $\tau$ increases from 0.5 to 0.9, the false-alarm rate decreases from 64\% to 14\%, but actionable detection also drops from 88\% to 33\%, while the median lead time decreases from 190 to 70 steps. These results highlight the trade-off between reducing false alarms and providing sufficiently early warnings.

    \subsection{Discovering Outcome-Associated SAE Features}

        Next, we seek SAE features whose activations may associate with successful or failed outcomes, potentially revealing specific success- or failure-related behavioral patterns.
        For feature $j$, we define the rolling mean activation as $m_{i,t,j}=\frac{1}{T}\sum_{s=t-T+1}^{t}a_{i,s,j}$, and the \textbf{trajectory score} as the maximum rolling mean activation, $s_{i,j}=\max_{t\geq T}m_{i,t,j}$.
        We hypothesize that some features exhibit strong indications of success or failure when highly activated. We therefore select the top-$k$ trajectories by trajectory score for each feature $j$ and compute their success rate among these top-$k$ trajectories. We use $k=30$ in our experiments.
        Table~\ref{tab:explanations-main} shows a selection of features with the lowest and highest success rates (see full results in Table~\ref{tab:sae-outcome-features}), which supports our hypothesis that strong activation is predictive of the trajectory outcome:
        trajectories with high scores for feature 1578 are almost always successful (28 of the top 30), whereas those with high scores for feature 480 are almost always failures (29 of the top 30, spanning four tasks).
        This suggests that individual SAE features may correspond to specific success- or failure-related behavioral patterns.
        \begin{table}[t]
        
            \centering
            \scriptsize
            \caption{
                Representative outcome-associated SAE features and
                their automatically inferred behavioral patterns.
                Success rates are based on each feature's top 30 trajectories.
                AUROC evaluates descriptions on held-out clips; $\Delta$ is the difference from shuffled-description AUROC. Task descriptions are provided in Table~\ref{tab:tasks}.
            }
            \label{tab:explanations-main}
            \renewcommand{\arraystretch}{1.1}
            \setlength{\tabcolsep}{4pt}
            \begin{tabularx}{\linewidth}{@{}rrXrr@{}}
                \toprule
                Feature & Success rate & Automatically inferred pattern
                & AUROC & $\Delta$ \\
                \midrule
                480 & 3.33\% &
                In \emph{Task 9}, arm moves right from the stove toward
                the moka pot handle without holding the pot
                & 0.72 & 0.26 \\
        
                467 & 3.33\% &
                In \emph{Task 1}, gripper hovers close to the moka pot
                handle without moving the pot
                & 0.84 & 0.22 \\
        
                850 & 3.33\% &
                In \emph{Task 1}, open gripper approaching the
                not-yet-grasped middle moka pot
                & 0.42 & $-$0.53 \\
        
                \midrule

        1578 & 93.33\% &
        In \emph{Task 4}, carrying the grasped task object left toward the basket
        & 0.89 & 0.04 \\
        1671 & 86.67\% &
        In \emph{Task 3}, carrying the grasped green-and-red-label
        can left and placing it in the basket
        & 1.00 & 0.17 \\

        \midrule
        \multicolumn{3}{@{}l}{
            All 12 feature--task pairs
            (261 clips; shuffled-description AUROC 0.58)
        }
        & 0.76 & 0.17 \\
        \bottomrule
    \end{tabularx}
    \vskip -0.3in
\end{table}
    
     \subsection{Localizing Outcome-Associated Behavioral Patterns}

\begin{figure*}[t]
            \vskip -0.1in
            \centering
            
            \includegraphics[width=0.7\textwidth]{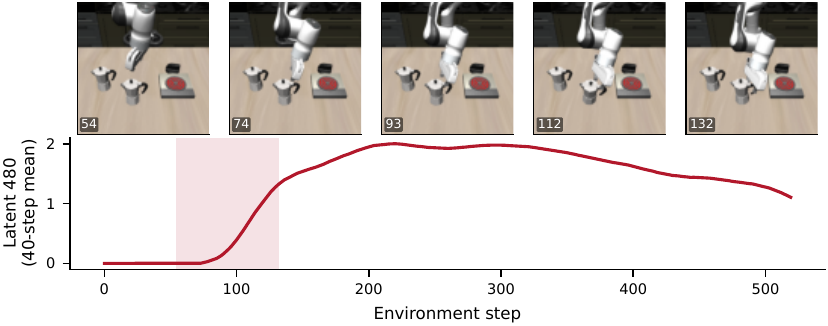}
            \vskip -0.1in
            \caption{Activation traces over time and automatically detected critical intervals for the failed trajectories with the strongest activations of
            feature 480 in Task~1: "Put both moka pots on the stove". It shows five uniformly
            sampled visualization frames above the full activation trace. 
            }
            
            \label{fig:feature480-critical}
            \vskip -0.20in
        \end{figure*}

        We first inspect the temporal activation patterns of outcome-associated SAE features to understand their behavioral significance. As shown in Figure~\ref{fig:feature480-critical}, feature 480's activation rises sharply and subsequently remains elevated. This pattern is consistent across its strongest-activating failed trajectories (see Figure~\ref{fig:feature480-critical-top3} in Appendix).
        Manual inspection reveals that high activation coincides with the robot remaining nearly stationary near the target object. 
        We observe similar associations for other features, with high activations corresponding to specific failure states in failed trajectories or particular stages of task execution in successful trajectories.

        Although high activation may indicate a failure state even before the episode terminates, we are more interested in the events leading into these states, as they may offer insights into potential causal mechanisms underlying success or failure and inform future policy improvements. We therefore detect the strongest sustained rise in each feature trace, rather than simply selecting regions of high activation, and retain one naturally bounded critical interval per trajectory (See details in Appendix~\ref{app:critical_interval}). Figure~\ref{fig:feature480-critical-top3} visualizes the automatically selected intervals for the three failure trajectories that have the strongest feature 480 activations in Task~1 (put both moka pots on the stove). Despite differences in timing, all three intervals show a similar pattern: the robot reaches the middle moka pot and remains in contact with it, but fails to transport it onto the stove. These observations suggest that the detected activation rises can consistently localize behavioral events that may contribute to failure-related states.

        \subsection{Automatic Interpretation and Validation}
    
        Manual inspection is useful for validation but does not scale.
        We therefore develop an automatic interpretation pipeline that uses a VLM to describe behavioral patterns in critical intervals (See details in Appendix~\ref{app:full_auto_interp}).
        Table~\ref{tab:explanations-main} presents two representative automatic interpretations: failure-related feature 480 in Task~9 (turn on the stove and put the moka pot on it) and success-related feature 1671 in Task~3 (put both the alphabet soup and the tomato sauce in the basket).
        Feature 480 in Task~9 most clearly captures a stalled approach to a moka pot that never moves with the gripper, whereas feature 1671 consistently captures transporting a grasped can into the basket.
        The explanations may point to events that contribute to the subsequent failure or success and thus provide candidate causal explanations for the final outcome.
        More interpretation results and analysis are provided in Table~\ref{tab:explanations} in Appendix~\ref{app:full_auto_interp}.
        We validate the generated descriptions using detection scoring~\cite{paulo2024automatically} on held-out clips. Across 12 feature--task pairs, descriptions achieve a mean
        AUROC of 0.76, compared to 0.58 for shuffled descriptions (Appendix~\ref{app:explanation_validation}).
        
\section{Conclusion}
\label{sec:conclusion}
    This work presents a novel framework that goes beyond early failure prediction to automatically discover and interpret outcome-associated behavioral patterns in VLA models. 
    We show that SAE features preserve outcome-predictive information in OpenVLA activations, enabling failure prediction before the robot reaches an estimated unrecoverable state.
    We further identify individual SAE features strongly associated with successful or failed outcomes and uncover recurring behavioral patterns by localizing critical activation intervals.
    Our automatic VLM-based interpretation pipeline generates behavioral descriptions of these patterns, which we validate against feature activations on held-out clips.
    Together, these results connect early failure prediction with interpretable analysis of outcome-associated behaviors, providing a foundation for investigating potential failure mechanisms, developing targeted safety interventions, and exploring interpretable features as reward signals for future policy training.


\clearpage
\acknowledgments{We want to thank Dr. Kenneth Marino and the students of the University of Utah's Spring 2026 CS 6960 course on multimodal llm agents for their feedback at the start of the project.}


\bibliography{main}  

\appendix
\section{Related Work}
\label{sec:citations-extended}
        
    \textbf{Failure Prediction for Vision-Language-Action Models\quad}
    As vision-language-action (VLA) models such as RT-2~\cite{zitkovich2023rt2}, OpenVLA~\cite{kim2024openvla}, and $\pi_0$~\cite{black2024pi0} increasingly serve as general-purpose robot policies, recent work has explored whether their internal representations can provide early signals of task failure.
    SAFE~\cite{gu2025safe} learns a failure detector directly from internal VLA features and demonstrates multitask failure detection, using conformal prediction to trade off detection accuracy and warning time.
    More recently, Mahato and Ren~\cite{mahato2026early} showed that OpenVLA feedforward activations contain linearly decodable signals of near-term failure under visual distribution shifts, demonstrating that internal activations can provide early warning signals.
    FPC-VLA~\cite{yang2026fpcvla} instead uses an external VLM-based supervisor at manipulation keyframes to predict failures and generate corrective actions.
    These works establish that failures can be anticipated from either internal VLA representations or external observations.
    Our work complements this direction by asking whether the internal signals predictive of trajectory outcomes can be decomposed into sparse, individually interpretable features that reveal the behavioral events associated with success and failure.
    Unlike these works, which report warning time relative to the end of an episode, we also measure it relative to a simulator-estimated point after which the failure can no longer be avoided, because a warning is not early if it arrives after the failure has become unavoidable.
    
    \textbf{Sparse and Mechanistic Interpretability of VLA Policies\quad}
    Sparse autoencoders (SAEs) have emerged as a tool for decomposing dense neural representations into sparse and potentially interpretable features~\cite{bricken2023towards,cunningham2023sparse}. 
    Recent work has extended these techniques to VLA policies. Swann et al.~\cite{swann2026sparse} train SAEs on VLA hidden activations and identify both episode-specific features and generalizable motion or semantic features, demonstrating through intervention that some individual SAE features can causally steer robot behavior. 
    Event-Grounded SAEs~\cite{jin2026eventgrounded} further ground SAE features in behaviorally salient events derived from robot trajectories, using visual, state, and temporal cues to associate internal features with closed-loop behavior and evaluate them through interventions. 
    These results demonstrate that SAE features can expose behaviorally meaningful structure within VLA representations. 
    Automatic feature descriptions are commonly validated by detection scoring, in which a model that sees only the description predicts held-out activations~\cite{bills2023language,paulo2024automatically}; Gur-Arieh et al.~\cite{gurarieh2025outputcentric} distinguish such input-centric descriptions from output-centric descriptions of a feature's effect on the model's output.
    In contrast, we specifically connect sparse features to trajectory outcomes: we first identify SAE features whose activation is strongly associated with success or failure, then localize critical temporal intervals in which these features emerge and use a VLM to automatically interpret the corresponding behavioral events. 

\section{Experimental Setup Details}
    \label{app:setup}
    \subsection{Dataset and Rollout Collection}
    We study OpenVLA~\cite{kim2024openvla} rollouts on multiple robot manipulation tasks. Each rollout terminates either when the task is successfully completed or when it reaches the maximum number of environment steps, in which case it is considered a failure. We use OpenVLA-7B fine-tuned on LIBERO-10~\cite{liu2023libero} and run it greedily from the 50 fixed benchmark initial states of each of the ten tasks (500 rollouts, at most 520 steps after 10 settling steps), recording each camera frame and the full simulator state. Overall, 51.6\% of rollouts succeed, ranging from 12\% to 80\% across tasks (Table~\ref{tab:tasks}; tasks are numbered alphabetically by instruction). A hash of the trajectory identifier selects 95 held-out trajectories (51 failures, 44 successes), and the remaining trajectories are split into 324 fit and 81 calibration trajectories; all results are reported on the held-out set unless stated otherwise.
        Let $\mathbf{h}_{i,t}$ denote the layer-0 residual-stream activation of OpenVLA at timestep $t$ of trajectory $i$, averaged over the image tokens. 
    
        \begin{table}[t]
            \centering
            \small
            \caption{LIBERO-10 tasks (numbered alphabetically by instruction) and the success rate of OpenVLA over the 50 benchmark initial states. }
            \label{tab:tasks}
            \begin{tabular}{rp{9cm}r}
                \toprule
                Task & Instruction & Success \\
                \midrule
                0 & pick up the book and place it in the back compartment of the caddy & 80\% \\
                1 & put both moka pots on the stove & 12\% \\
                2 & put both the alphabet soup and the cream cheese box in the basket & 56\% \\
                3 & put both the alphabet soup and the tomato sauce in the basket & 54\% \\
                4 & put both the cream cheese box and the butter in the basket & 76\% \\
                5 & put the black bowl in the bottom drawer of the cabinet and close it & 46\% \\
                6 & put the white mug on the left plate and put the yellow and white mug on the right plate & 50\% \\
                7 & put the white mug on the plate and put the chocolate pudding to the right of the plate & 44\% \\
                8 & put the yellow and white mug in the microwave and close it & 36\% \\
                9 & turn on the stove and put the moka pot on it & 62\% \\
                \bottomrule
            \end{tabular}
        \end{table}

    \subsection{Sparse Autoencoder}
    We aim to understand the behavioral patterns encoded in these activations, but dense representations can be difficult to interpret because individual dimensions may represent multiple features due to superposition and polysemanticity~\cite{bricken2023towards,cunningham2023sparse}. We therefore use a top-K sparse autoencoder (SAE) to decompose $\mathbf{h}_{i,t}$ into sparse feature activations $\mathbf{a}_{i,t}$. 
    Specifically, the SAE encoder maps $\mathbf{h}_{i,t}$ to feature activations
    $\mathbf{z}_{i,t}
    =
    \operatorname{ReLU}
    \left(
    \mathbf{W}_{\mathrm{enc}}
    \left(
    \mathbf{h}_{i,t}-\mathbf{b}_{\mathrm{dec}}
    \right)
    +
    \mathbf{b}_{\mathrm{enc}}
    \right)$.
    We retain only the top-$K$ activations to obtain the sparse representation
    $\mathbf{a}_{i,t}
    =
    \operatorname{TopK}
    \left(
    \mathbf{z}_{i,t},K
    \right)$,
    and reconstruct the original activation as
    $\hat{\mathbf{h}}_{i,t}
    =
    \mathbf{W}_{\mathrm{dec}}\mathbf{a}_{i,t}
    +
    \mathbf{b}_{\mathrm{dec}}$.
    This associates each feature with a specific activation direction that may correspond to a particular behavioral pattern, providing a more interpretable unit of analysis.
    We use $2{,}048$ features and $K=64$, and train the SAE on mean-centered activations from all 500 rollouts.
    Our goal is to determine whether $\mathbf{h}_{i,t}$ and $\mathbf{a}_{i,t}$ contain outcome-associated information, and whether individual SAE features reveal interpretable success- and failure-related behavioral patterns.

    \section{Additional Experimental Results}
    \label{app:additional_results}
    \begin{figure}[t]
            \centering
            \begin{subfigure}{\linewidth}
                \centering
                \includegraphics[width=\linewidth]{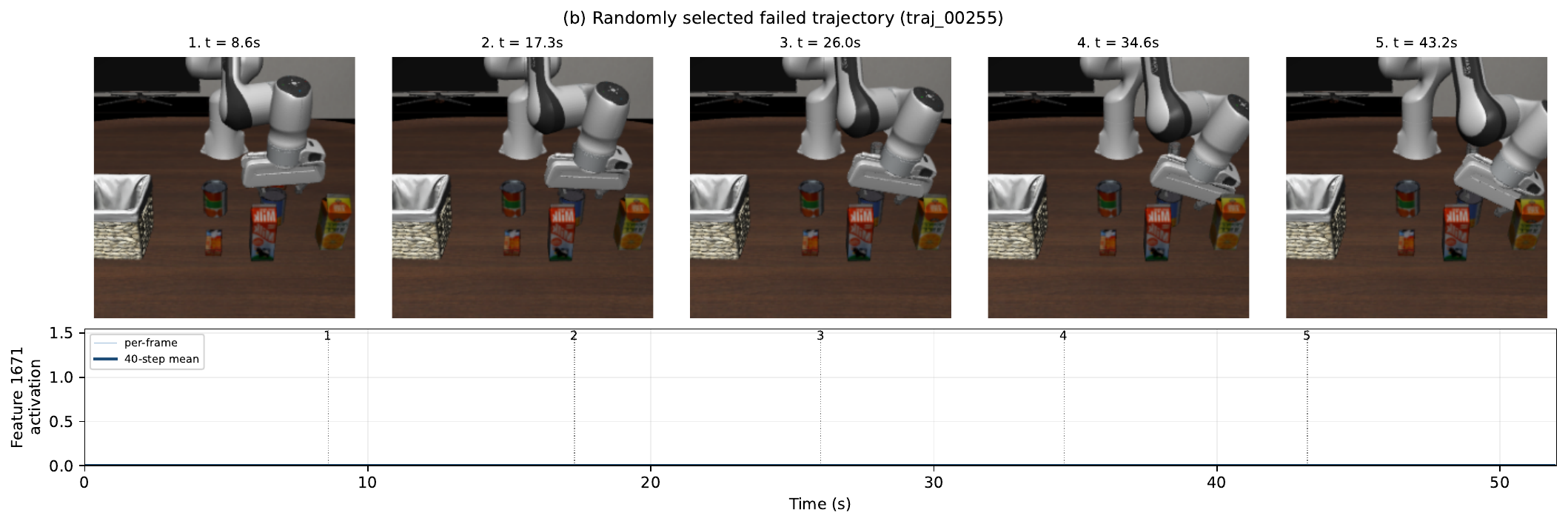}
                \caption{Randomly selected failed trajectory.}
            \end{subfigure}
            \vspace{0.6em}
            \begin{subfigure}{\linewidth}
                \centering
                \includegraphics[width=\linewidth]{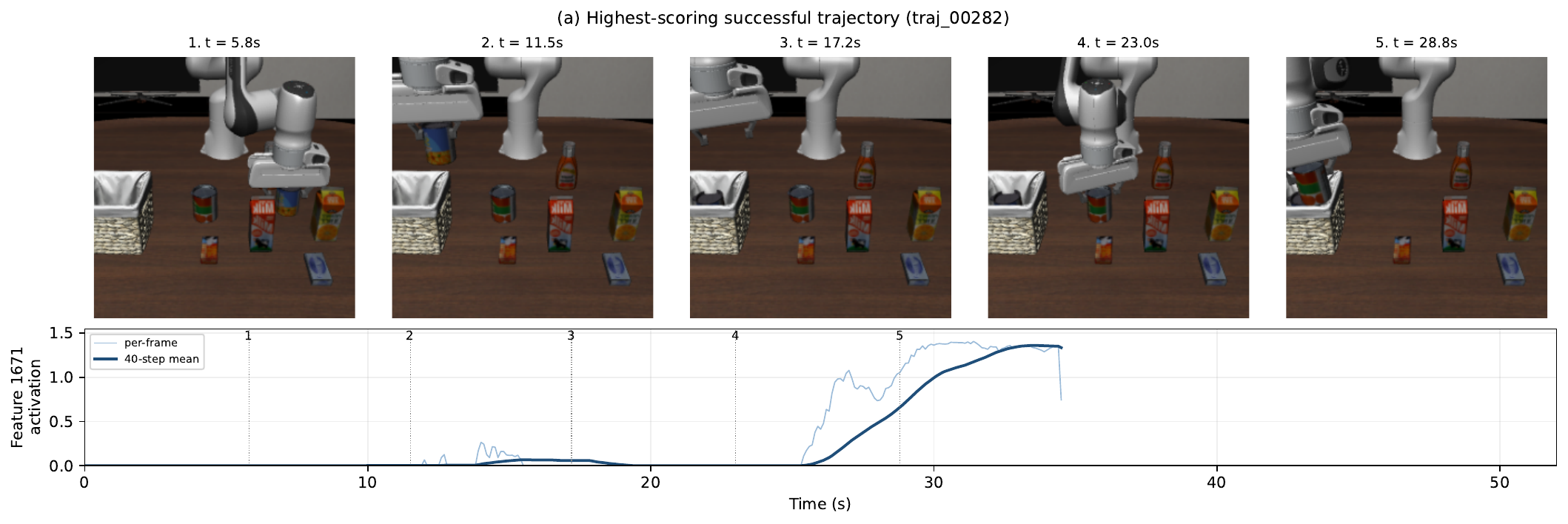}
                \caption{High-activation successful trajectory.}
            \end{subfigure}
            \caption{Temporal activation of success-related feature 1671 on Task~3.}
            \label{fig:feature1671-comparison}
        \end{figure}

      \begin{figure*}[t]
            \centering
            \includegraphics[width=1.0\textwidth]{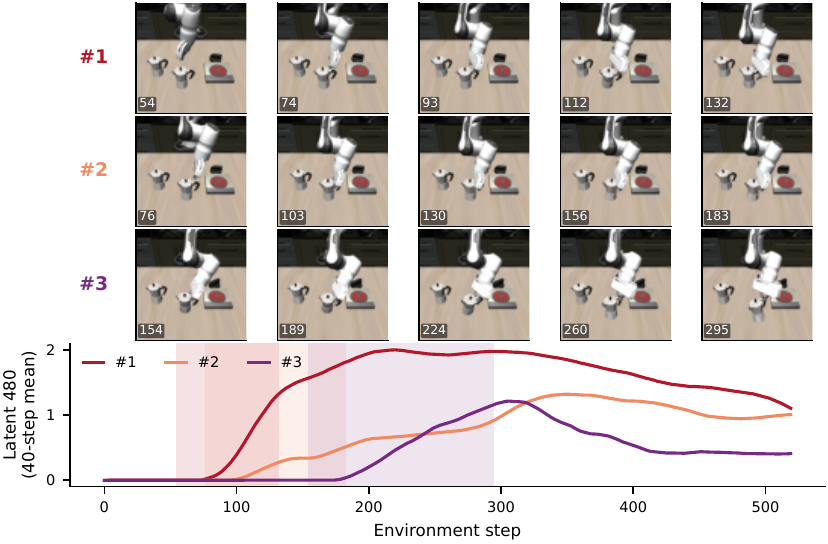}
            \caption{Activation traces over time and automatically detected critical intervals for three failed trajectories with the strongest activations of
            feature 480 in Task~1. It shows five uniformly
            sampled visualization frames above the full activation trace. 
            }
            \label{fig:feature480-critical-top3}
        \end{figure*}

    \subsection{Full Automatic Interpretation Results}
    \label{app:full_auto_interp}
    Table~\ref{tab:explanations} summarizes the automatic explanations for the four highest-ranked failure-related features (480, 850, 35, and 467) and the four highest-ranked success-related features (1578, 748, 132, and 1671) across every task with at least three eligible primary examples.
    The results reveal both useful task-specific semantics and important limitations. 
    Feature 480 in Tasks~1 and~9 most clearly captures a stalled approach to a moka pot that never moves with the gripper, whereas feature 1671 consistently captures transporting a grasped can into the basket.
    The explanations may point to an event that contributes to the subsequent failure state and thus provides a candidate causal explanation for failure.
    In contrast, feature 132 represents the final book-placement stage, while feature 480 in Task~8 is broader, describing the mug held near the open microwave rather than a stalled grasp, which shows that a feature's meaning can depend on the task.
    These features are better interpreted as markers of successful or failed outcomes rather than their causes.

    \begin{table}[t]
            \centering
            \caption{SAE features with the strongest outcome associations. For each feature, the success rate is computed over the top-$k$ trajectories. We show features with the lowest and highest success rates, corresponding to failure- and success-associated features, respectively.}
            \label{tab:sae-outcome-features}
            \begin{tabular}{llrlr}
                \toprule
                Relation & Feature & Success rate & \# Unique Tasks & Active trajectories \\
                \midrule
                    Failure & 480 & 3.33\% & 4 & 303 \\
                    Failure & 850 & 3.33\% & 1 & 164 \\
                    Failure & 35 & 3.33\% & 1 & 346 \\
                    Failure & 467 & 3.33\% & 1 & 240 \\
                    Failure & 685 & 6.67\% & 5 & 367 \\
                \midrule
                    Success & 1578 & 93.33\% & 3 & 235 \\
                    Success & 748 & 90.00\% & 1 & 176 \\
                    Success & 132 & 90.00\% & 1 & 132 \\
                    Success & 1671 & 86.67\% & 2 & 188 \\
                    Success & 1528 & 86.67\% & 1 & 230 \\
                \bottomrule
            \end{tabular}
        \end{table}

    \subsection{Explanation Validations}
    \label{app:explanation_validation}
     Table~\ref{tab:explanations} lists the per-pair scores. Across the 12 pairs (261 clips), the mean AUROC is 0.76 (95\% CI [0.64, 0.87], bootstrap over pairs and clips; permutation $p<0.001$) and the mean Spearman correlation with activation is 0.46. Descriptions outperform shuffled descriptions on 11 of 12 pairs (mean $\Delta$AUROC $=0.17$, 95\% CI over clips [0.10, 0.24]; sign test $p=0.003$), although resampling pairs as well widens this interval to [$-0.01$, 0.35], so specificity varies across features. Individual descriptions are uneven: the description of feature 850 is anti-correlated with its activation (AUROC 0.42), and several transport descriptions are accurate but not specific, because other features on the same task describe similar transport (e.g.\ feature 132: 0.97 versus 0.92 shuffled). With 24 clips per pair, an individual AUROC has a confidence interval of roughly $\pm0.2$; we therefore treat the description methodology as validated in aggregate and individual descriptions as hypotheses.
    
    \begin{table*}[t]
            \centering
            \small
            \caption{Automatic interpretations of the four highest-ranked failure-related and success-related SAE features and their reproduction-test scores on held-out clips: AUROC of a judge VLM that sees only the description, and the difference $\Delta$ to a shuffled description of another feature on the same clips. Each explanation uses the three strongest task-specific critical clips and one independently sampled successful demonstration. The outcome-relation category is produced by the external VLM. Task numbers follow Table~\ref{tab:tasks}.}
            \label{tab:explanations}
            \renewcommand{\arraystretch}{1.12}
            \begin{tabularx}{\textwidth}{llcXlrr}
                \toprule
                    Relation & Feature & Task & Automatically inferred pattern & Outcome relation & AUROC & $\Delta$ \\
                \midrule
                    Failure & 35 & 1 & Empty arm swings right into the workspace before touching any object & ordinary task progress & 0.53 & 0.46 \\
                    Failure & 467 & 1 & Gripper hovers close to the moka pot handle without moving the pot & unclear & 0.84 & 0.22 \\
                    Failure & 480 & 1 & Stalled, unlifted, or pushing grasp attempt after closing in on the moka pot & early indicator & 0.56 & 0.06 \\
                    Failure & 480 & 8 & End effector holding or touching the yellow-and-white mug and moving it toward the open microwave & ordinary task progress & 0.65 & 0.15 \\
                    Failure & 480 & 9 & Arm moves right from the stove toward the moka pot handle / frying pan area without holding the pot & early indicator & 0.72 & 0.26 \\
                    Failure & 850 & 1 & Open gripper approaching the not-yet-grasped middle moka pot & ordinary task progress & 0.42 & $-$0.53 \\
                    Success & 132 & 0 & Placing the grasped book into the back compartment of the caddy and releasing it & ordinary task progress & 0.97 & 0.05 \\
                    Success & 748 & 0 & Lifting the grasped book and carrying it toward the back compartment & ordinary task progress & 0.88 & 0.65 \\
                    Success & 1578 & 3 & Carrying the grasped blue-and-yellow can left to the basket and placing it inside & ordinary task progress & 0.78 & 0.06 \\
                    Success & 1578 & 4 & Carrying the grasped task object left toward the basket & ordinary task progress & 0.89 & 0.04 \\
                    Success & 1671 & 2 & Delivering the grasped blue-and-yellow can into the basket (can at or inside the basket opening) & ordinary task progress & 0.86 & 0.50 \\
                    Success & 1671 & 3 & Carrying the grasped green-and-red-label can left and placing it in the basket & ordinary task progress & 1.00 & 0.17 \\
                \bottomrule
            \end{tabularx}
        \end{table*}

    \section{Selection of Critical Activation Intervals}
    \label{app:critical_interval}
    Given the activation sequence $\{a_t\}_{t=1}^{T}$ of an SAE feature along a
    trajectory, we identify one critical activation interval that captures the most
    salient transition into a strongly activated state. This is intended to exclude
    isolated activation spikes and to retain sufficient visual context before and
    after the transition.
    
    We first normalize the activation sequence by its 90th percentile over non-zero
    activations and apply Gaussian smoothing. For each time step $t$, we then compute
    a local activation contrast:
    \[
    c_t =
    \operatorname{median}
    \left(\tilde{a}_{t:t+h}\right)
    -
    \operatorname{median}
    \left(\tilde{a}_{t-h:t}\right),
    \]
    where $\tilde{a}_t$ is the smoothed normalized activation and $h$ is a temporal
    horizon. In our experiments, $h=15$ environment steps. A high value of $c_t$
    therefore indicates that the feature becomes consistently more active after
    $t$, rather than exhibiting only a single-frame spike.
    
    Candidate rising events are obtained from prominent local maxima of $c_t$. The
    boundaries of each event are determined by extending from its peak while the
    local contrast remains above a relative support threshold. We additionally
    include 15 steps of temporal context on both sides of the detected transition,
    so that the resulting clip contains part of the preceding low-activation state,
    the activation rise, and the subsequent high-activation state.
    
    Nearby candidate events separated by fewer than 25 steps are merged. This makes
    the detector robust to short activation drops or intermittent zero activations
    within the same underlying event. If multiple event groups remain, we retain
    the group with the largest aggregate prominence. Thus, exactly one critical
    interval is selected for each trajectory.
    All detector parameters are fixed across features, tasks, and trajectory
    outcomes. Outcome labels are not used during interval selection.

        \section{Recoverability Estimation}
    \label{app:recoverability}
    This appendix details the recoverability estimate and point of no return (PNR) used to evaluate early warning in Section~\ref{sec:online-prediction}.
 
    \subsection{Recoverability Estimator}
    \label{app:recoverability-estimator}
    OpenVLA conditions only on the current frame and instruction, so the recorded simulator state $s_t$ determines the policy's future. We estimate the recoverability $R(s_t)=\Pr[\text{success}\mid s_t]$ under the policy with temperature-1 sampling of its action tokens by restoring $s_t$ and running $B=4$ stochastic continuations for the remaining $520-t$ steps:
    \[
    \hat R_t=\frac{1}{B}\sum_{b=1}^{B}\mathbb{1}[\text{branch } b \text{ succeeds}].
    \]
    Sampling is necessary because the recorded rollouts are collected with greedy decoding (Appendix~\ref{app:setup}), which would reproduce the recorded outcome.
 
    \subsection{Probe Placement and Point of No Return}
    \label{app:recoverability-pnr}
    Probes are placed at $-40,\dots,+40$ steps around the predictor's alarm and around abrupt rises in its failure log-odds in all 51 held-out failures (838 probes, 3{,}509 branch rollouts). For a failed trajectory with probes $t_1<\dots<t_n$, the PNR is the earliest probe after which every probe has $\hat R\le0.1$; it is left-censored if this already holds at the first probe. The predictor raises an \emph{alarm} once $p_{\mathrm{fail}}\ge\tau$ for two consecutive evaluations, and an alarm is \emph{actionable} if it precedes the PNR.
 
    For the before-PNR evaluation in Section~\ref{sec:online-prediction}, each failure is restricted to frames up to its last still-recoverable probe and compared with held-out successful trajectories at matched times. This uses the 25 failures with an uncensored PNR before step 420; later PNRs are excluded because late probes can fail simply because the remaining step budget runs out (Appendix~\ref{app:recoverability-limitations}).
 
    \subsection{Short-Term Spikes in Predicted Failure Probability}
    \label{app:recoverability-spikes}
    Short-term spikes in $p_{\mathrm{fail}}$ are less informative than its gradual rise: most spikes (83\%) occur after $\hat R$ has already fallen to at most 0.1, and those occurring earlier show no larger drop in $\hat R$ than matched non-spike intervals (difference $-0.02$, 95\% CI [$-0.12$, 0.08]). With four branches per probe, smaller effects cannot be excluded. We also note that our methodology for detecting spikes, qualitatively from human analysis, seems to exclude important meso-trends over the course of tens of timesteps that predict human-derived PNR values. We leave extending this pipeline to account for this confound to future analysis.
 
    \subsection{Limitations of the Recoverability Estimate}
    \label{app:recoverability-limitations}
    $\hat R$ is an imperfect instrument. Branches inherit the remaining step budget, so late probes can fail because time runs out (branch success is 0.2\% with fewer than 100 steps left versus 18-23\% with 200, regardless of actual simulator state), which is why we exclude PNRs after step 420. Temperature sampling perturbs the policy in a phase-dependent way: in one rollout the greedy policy was stuck against the microwave roof from about step 207, but sampled branches occasionally shook free and delayed the measured PNR, while during fast motions or object transport sampled branches drop objects more often than the greedy policy, which lowers $\hat R$. With four branches per probe a single estimate is coarse (for $\hat R=0.25$ the 95\% interval is roughly [0.05, 0.70]), and $\hat R$ measures recoverability for this policy rather than physical recoverability. We therefore read these results as estimates that support qualitative conclusions, not precise failure times. We hope to extend this methodology in the future to involve more branches and further development of the recoverability estimation system.

\end{document}